# APPLICATION OF THE YOLOV5 MODEL FOR THE DETECTION OF MICRO-OBJECTS IN THE MARINE ENVIRONMENT

A.N. Grekov[1, 2], Y.E. Shishkin[1], S.S. Peliushenko[1], A.S. Mavrin[1, 2]

[1]Institute of Natural and Technical Systems, RF, Sevastopol, Lenin St., 28
[2]Sevastopol State University, RF, Sevastopol, Universitetskaya St., 33

The efficiency of using the YOLOV5 machine learning model for solving the problem of automatic detection and recognition of micro-objects in the marine environment is studied. Samples of microplankton and microplastics were prepared, according to which a database of classified images was collected for training an image recognition neural network. The results of experiments using a trained network to find micro-objects in photo and video images in real time are presented. Experimental studies have shown high efficiency, comparable to manual recognition, of the proposed model in solving problems of detecting micro-objects in the marine environment.

**Introduction.** In modern science, machine vision is one of the promising methods for studying the marine environment. Over the past decade, great progress has been made in the field of object detection on video images in real time. Object detection is considered as a computer vision method that is used to find and identify objects [1-4].

Traditional object detection methods usually consist of 3 stages [2]:

1) determining the location of objects. To do this, the image is scanned using a sliding window, which is computationally expensive and may produce irrelevant candidates;

2) feature extraction using the methods HOG (Histogram of Oriented Gradient) [3], SIFT (Scale Invariant Feature Transform) [4], etc. However, due to different lighting conditions, changes in angles and backgrounds, it is very difficult to manually develop a reliable object descriptor to accurately describe all of their types;

3) the use of specialized classifiers (for example, the SVM classifier) for object recognition.

The use of new approaches of deep learning (DL) has significantly improved the quality of solving the problem of machine vision due to the implementation of hardware calculations on GPUs for image classification and object detection [5]. After the significantly increased performance of computing equipment, deep neural networks (DNNs - Deep neural networks) have become widely used in the field of detecting objects, such as plankton, due to their ability to extract features more efficiently than traditional methods. [6, 7]. GO-based object detection algorithms fall into two categories: two-stage methods and one-stage methods. In 2014, Ross Grishik et al. [8] proposed a two-stage R-CNN detection algorithm (Regionswith CNN Features), and were the first to apply a convolutional neural network (CNN) to the problem of object detection. The study showed an almost 30% improvement in object detection rate compared to earlier methods. This selective search approach for extracting regions of an image consists of two steps: the first step is to extract the possible features of an object using a region-based method, and the second step is to extract features from each region using a CNN.

Two-stage discovery algorithms such as Fast RCNN [9] and Faster RCNN [10] appeared somewhat later and are a development of HOG and SIFT. First, an updated version of Fast RCNN [11] was published, which uses selective search to generate var-ious objects, and then Faster RCNN. Instead of extracting them all independently using SVM classifiers, the new approach applies a CNN to the entire image. Both regions of interest (RoIs) are then used, combining the feature map with the final feedforward network for classification and regression. Compared to previous works on deep learning, Fast R-CNN has a higher learning rate and detection accuracy. However, two-stage methods have greater computational complexity and low inference speed due to the large number of prediction blocks, as a result of which their use for embedded platforms is very limited.

A large number of studies are aimed at the use of universal neural networks for solving highly specialized tasks, for example, for recognizing plankton. AlexNet and VGGNet, Dai J et al. [6] proposed a problem-oriented neural network for recognizing microplankton ZooplanktoNet, consisting of 11 layers and achieving an accuracy (correctness metric) of 93.7%. Lee H et al. [12] and Py O et al. [13] used a deep residual network and a deep CNN with a multidimensional image recognition engine to classify plankton. Lee et al. [14, 15] enabled transfer learning by pre-training a CNN with class-normalized data and fine-tuning with the original data on an open dataset called WHOI Plankton. It is shown that the accuracy (correctness metric) of the classification increases, but there remains a serious problem with the quality of prediction for rare anomalous classes. Lumini A. et al. have been working on fine-tuning and transfer learning of several well-known deep learning models (AlexNet, GoogleNet, VGG, etc.) to develop an ensemble of plankton classifiers. The performance of this approach surpassed earlier models, and the level of recognition accuracy (correctness metric) reached 95.3% in an experiment with a set of WHOI-Plankton database images.

Shaoking Ren et al. [16], instead of using selective search, as in R-CNN and Fast R-CNN, used region proposal networks (RPN) technology to extract image regions. RPN makes full use of image convolutional functions that predict object boundaries and objectless estimates of each position. Fast R-CNNs and RPNs were also combined to form a single unified network based on deep learning using a common convolution function.

Wei Liu et al. [17] presented a single DNN based object detection method called the Single ShotMultiBox Detector (SSD). The new approach is based on a feedforward multi-scale convolutional network that shares the output space of the target blocks with a different aspect ratio for the input image. The SSD method is easy to learn and applicable for integration with systems that require discovery. Experimental results show higher performance and accuracy if the resolution of the original images is low.

A promising method for detecting micro-objects in the marine environment is the use of one-stage models such as YOLO. Redmon et al. [18] presented a new unified approach to object detection called You Only Look Once (YOLO). YOLO uses one neural network for the entire image in determining the probability of objects belonging to classes and constructing bounding boxes using a single assessment. This method allows you to implement very fast real-time object detectors and process the image at a speed of 155 frames per second. Redmon et al. showed that compared to alternative models, YOLO has a lower probability of false predictions in an image. In the course of further development, such models as YOLOv2 [19], YOLOv3 [20], SSD [21] and other one-stage discovery networks appeared, which have a higher discovery rate sufficient to meet the requirements of work in real time.

YOLO is considered one of the best object detection models. Many researchers have compared it with other models for evaluation, and successfully use it in their work. In the work of Tamoshkin M.S. et al. [22] compared YOLOv5, Faster R-CNN, etc., where it was confirmed that YOLOv5 works much faster than others, while losing 2-7% in accuracy (correctness metric).

Filichkin S.A. and Vologdin S.V. [23] used YOLOv5 to recognize the presence of personal protective equipment in an image,

and recognized the high efficiency of this model for such tasks.

Liang T.Zh. et al. [24] proposed an improved detection method in the image of a vehicle wheel weld based on the YOLOv4 object detection model and confirmed its effectiveness in this area of research.

Kaplunenko D.D. et al. [25] studied the possibility of using the YOLOv5 model to classify underwater living objects that dynamically and independently move in the field of view of underwater cameras of a stationary installation. In the course of the work, it was demonstrated that the model allows solving problems of this type, however, an acute dependence of the quality of the solution on the completeness of the data set used to train the neural network was revealed.

Also, the YOLOv5 algorithm is used to detect objects in problems of monitoring marine phytoplankton. For example, A. Pedraza et al. [26] applied the RCNN and YOLO object detection models to 10 classes of microscopic images of diatoms. Yu. Li et al. [27] developed the DenseYOLOv3 network to improve the mAP (Mean Average Precision) metrics selected from eight categories of phytoplankton in the WHOI-Plankton dataset. To solve the problem of an unbalanced phytoplankton dataset, Y. Li et al. [28] applied a data synthesis technique using CycleGAN to improve detection efficiency. W. Li et al. [29] created a PMID2019 dataset containing 10,000 microscopic images of phytoplankton from 24 different categories and compared the performance of object detection algorithms, including Faster-RCNN, SSD, YOLOv3, and RetinaNet [30], at the same time, YOLOv3 turned out to be the leader in terms of speed. Shi Z. [31] used a modified YOLOV2 model to detect zooplankton in holographic images.

The work proposed by the authors considers a method for solving the problem of detecting and classifying microorganisms and microplastics in the marine environment using the YOLOV5 model. For this, samples of microplankton and microplastics of the marine environment were collected, photo and video materials were prepared on their basis for training the YOLOV5 neural network, the neural network was trained and tested in real time.

**Materials and methods.** In the presented work, the YOLOvX model was used, implemented in Python in the form of a freely distributed YOLOv5 library [18]. For labeling images, the freely distributed tool Label-Studio [32] was used.

The YOLO model is a set of algorithms where object detection problems are solved. YOLO is a one-stage deep learning algorithm that uses convolutional neural networks for object detection. YOLOv5 is a family of object detection models that includes YOLO-based machine vision best practices and techniques compiled by author Glenn Joher based on previous versions of YOLOv1-YOLOv4, authored by Joseph Redmon, Ali Farhadi , Alexei Bochoknovsky, Jian-Yao Wang and Hong-Yuan Mark Liao.

Among the advantages of using YOLOv5 for embedded systems for monitoring the aquatic environment, it is worth highlighting the following:

– a small amount of memory is required to deploy the model;

– YOLO family models, in particular YOLOv5, are exceptionally fast and far superior to R-CNN and other models;

– the model distinguishes between the target and the background area better than others.

The general course of the study consisted of the following stages:

1) preparation of equipment required for work: computer, camera, sea water with microplastics and microorganisms;

2) video capture using a camera of a number of samples of images of water with microplastics and microorganisms. Within the framework of this work, 168 images were used for the training array, and 35 images were used for the control sample.

3) marking on images of the training sample using label-studio for the training array of the neural network is shown in (Fig. 1);

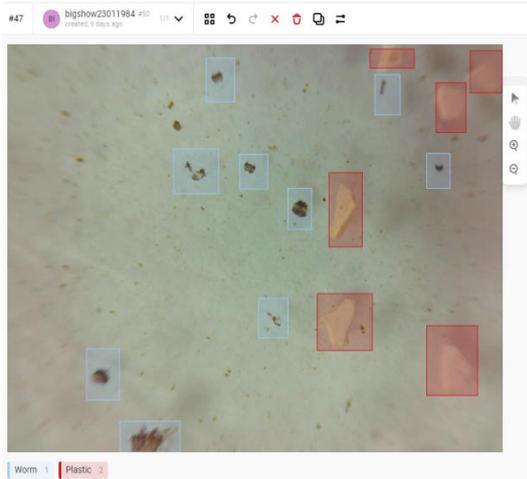

**Fig. 1.** Markup execution in label studio

4) export of the training array from label-studio in the format required for YOLOv5;

5) import of the training array into YOLOv5;

6) neural network training;

7) Using a neural network to detect and classify objects in the images of the control sample.

8) validation of the model according to the control sample;

9) comparison and analysis of the results.

The training array for use in the YOLOv5 model must have a certain structure, represented as a set of files divided into 4 groups:

1. images used to form the training sample;

2. data on the markup of the corresponding images (classes and coordinates of objects, dimensions of images);

3. information about all considered classes of objects.

4. description of classes of objects and the array of images of the training sample itself.

To obtain images, an ELP 5-50 digital camera was used, with a resolution of 3264x2448 pixels, a telephoto lens of 5-50 mm. The images were taken under laboratory conditions with illumination of 500 lux. Recognizable objects are divided into two classes: microplastics and microplankton, having geometric dimensions of 1-7 mm.

**Results.** To perform the work, a training sample array containing 168 images was prepared and used to train the neural network. The network is configured as follows: 50% of all images were used for neural network training, 25% for testing, 25% for validation. To study the efficiency of the model for different training periods, the network was trained on the same training array with a different number of iterations. For comparison, in Fig. 2(a) and fig. Figure 2(b) shows the results of detection and classification of objects in the studied image for a neural network trained on 50 and 500 iterations, respectively. In the visualization, a bounding box is displayed around the detected objects, allowing you to determine their position and movement in the image.

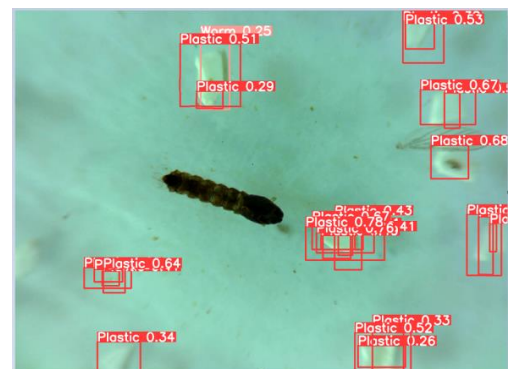

(а)

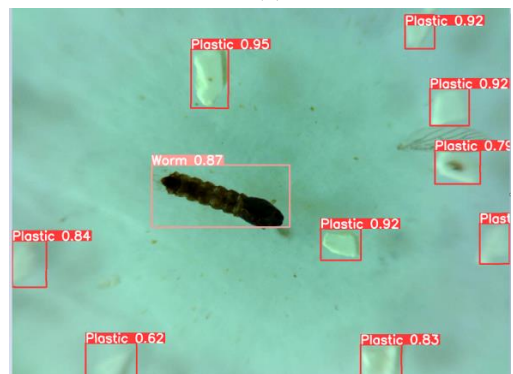

(б)

**Fig. 2.** Result for 50 (a), 500 iterations (b)

As the results of the experiments showed, with 500 iterations, all objects in the image were identified and correctly classified. At 50 iterations, all microplastics and microor-ganisms were also identified, however, there were a large number of false positives and duplicates.

To confirm the efficiency of the network operation at different training periods, numerical estimates of the neural network operation were compared, i.e. loss function metrics (regression loss, confidence loss, classification loss).

The essence of the loss function is to obtain a numerical estimate that determines how well the model performs. If the forecast is significantly different from the actual value, the numerical measure of the loss function will be very high. However, if both values are almost the same, the loss value will be very low. The loss function in YO-LO has three parts:

1. box_loss - loss of the bounding box. If the grid cell contains objects, then the loss of the bounding box will be calculated, i.e. how completely the object fits within the bounding box. With possible values of 0 at 100% object detection, and increases as the percentage falls.

2. obj_loss - loss of reliability. Specifies how accurately objects are separated from the image background.

3. clc_loss - loss of classification. Determines how correctly objects are distributed among classes.

These loss estimation functions are used to tune the model during training and to validate it. Graphs of the convergence dynamics of the loss estimation functions for 50 and 500 iterations are shown in fig. 3 and fig. 4 respectively.

On fig. Figure 5 shows an example of a successfully recognized image subjected to optical distortions typical of in situ underwater conditions: blurring, chromatic aberrations, scattering and refraction of light as it passes through an inhomogeneous air-water-glass medium. From the presented recognition results, it can be seen that in the presence of distortions, the accuracy of object recognition decreases with distance from the center of the image.

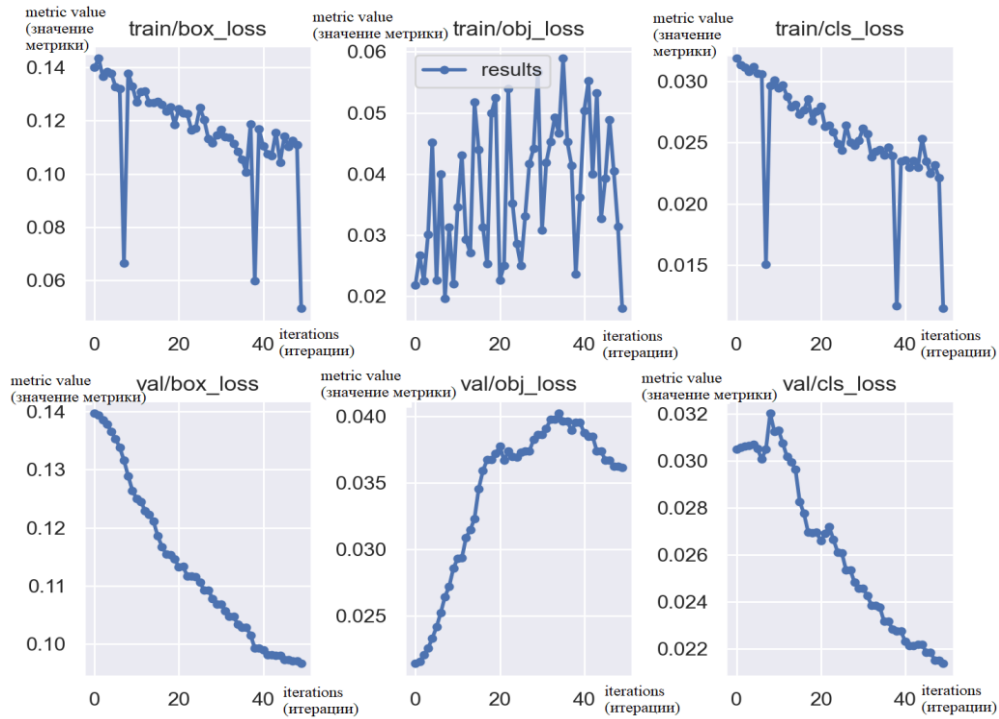

**Fig. 3.** Graphs of losses during training and recognition with a neural network for 50 iterations

As shown in fig. 4, with an increase in the number of iterations, a non-linear improvement in quality metrics occurs (less is better), which, for the collected database of plankton and microplastic images, reach values acceptable for practical use at 400 or more iterations.

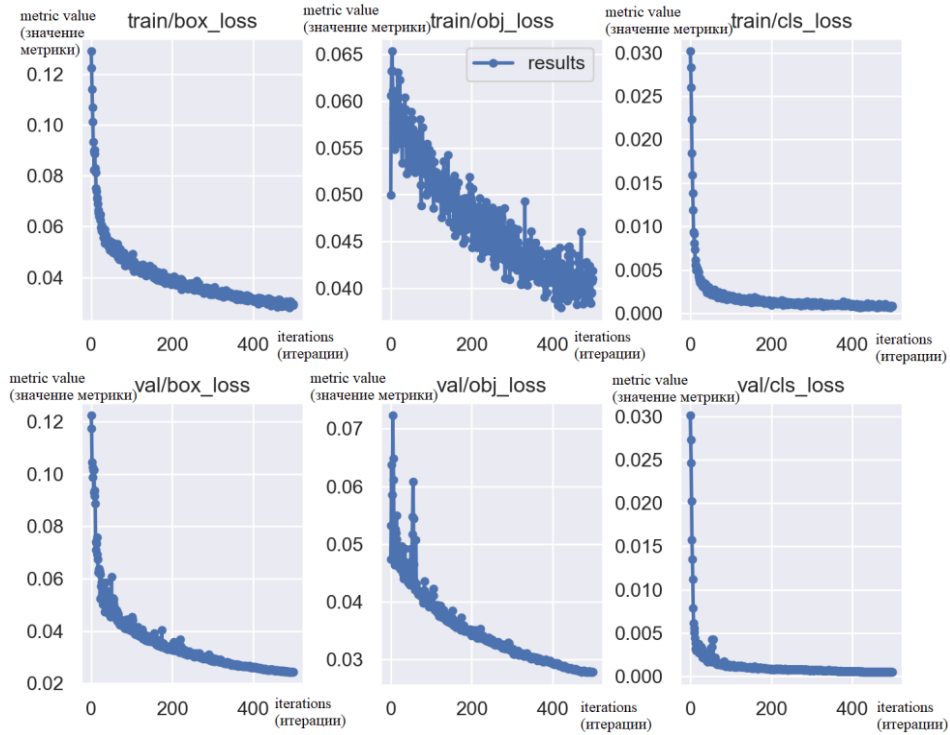

**Fig. 4.** Graphs of losses during training and recognition with a neural network for 500 iterations

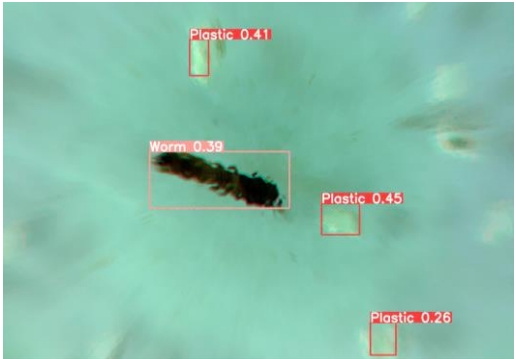

**Fig. 5.** Recognition results in the presence of optical distortions for 500 iterations

Next, the evaluation of the success of recognition by the neural network of the control sample by metrics was carried out:

1. precision - measures how accurate the predictions are, i.e. which part of the predictions is correct is described by formula (1).

$$precision = \frac{TP}{TP+FP} \quad (1)$$

where TP (True Positive) is the correct definition of the class, FP (False Positive) is the acceptance of the wrong object for the class.

2. recall - measures how well the desired objects are detected, described by formula (2).

$$recall = \frac{TP}{TP+FN} \quad (2)$$

where FN (FalseNegative) is non-selection of the class object.

3. mAP is the average value of AP for IoU = 0.5 (Intersection over union, measures the overlap between 2 boundaries, is used to measure how much the predicted boundary matches the real boundary of the object). It is one of the main performance metrics of a neural network. Average Precision calculates the average precision for recall between 0 and 1. IoU is used to determine the percentage of overlap between the predicted location of an object and its actual location. For some datasets, a threshold is predefined for the IoU and this is used to classify the object (eg, true positives).

4. mAP (0.5, 0.95) – average AP for IoU from 0.5 to 0.95, with a step size of 0.05.

These metrics for 50 and 500 iterations are presented in fig. 6 and fig. 7 respectively.

Integral recognition quality indicators (diagram of precision versus recall) for 50 and 500 iterations are shown in fig. 8(a) and fig. 8(b), respectively.

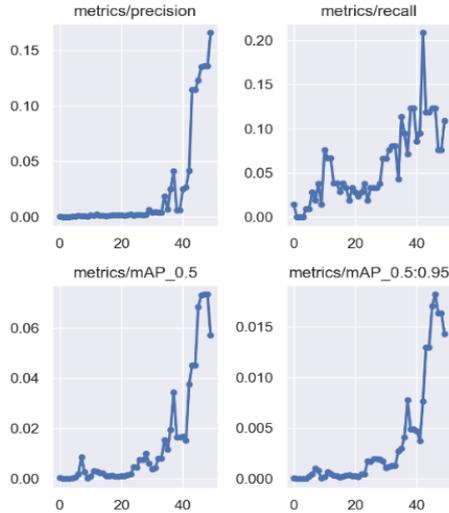

**Fig. 6.** Graphs of metrics for a neural network for 50 iterations

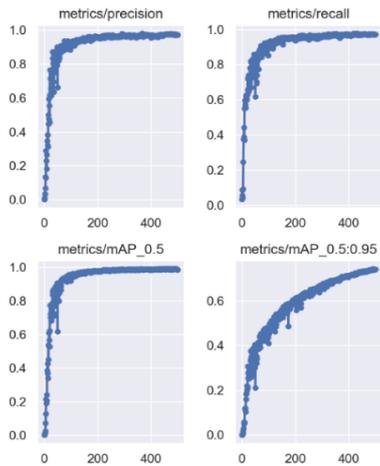

**Fig.7.** Graphs of metrics for a neural network for 500 iterations

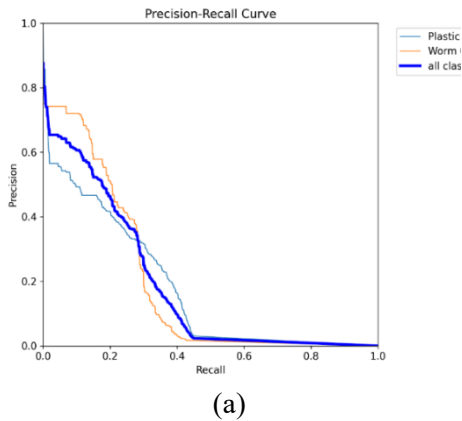

(a)

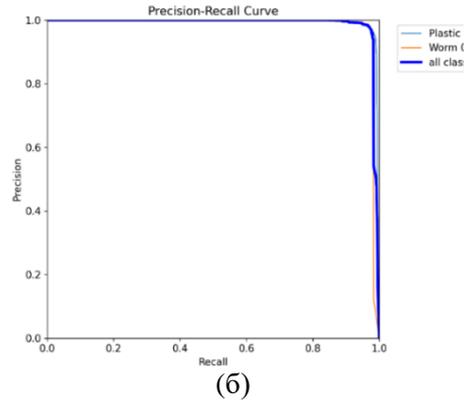

(б)

**Fig.8.** Precision-recall curve for a neural network for 50 epochs (a) and 500 epochs (б)

**Discussion and conclusion.** When considering the numerical metrics of the quality of the obtained model, the following can be noted:

1. As the number of iterations increases during training, the value of precision gradually grows, tending to 1. The program improves with the advancement of iterations, and begins to determine objects with almost 100% accuracy. On the considered example, the program reaches this level by 200 iterations. This means that the model correctly classifies objects in the image.

2. The recall value grows with the number of iterations, tending to 1. In the program, as the number of iterations increases, the completeness gradually increases, almost 100% of the required objects are found. This level in the program is reached by the 500th iteration.

3. The value of mAP also increases as the number of iterations increases up to 300, reaching almost 100% accuracy of the object detector. This means that this model detects objects in the image completely.

4. Losses show how well the model performs in the context of the presence of type II errors in detection and classification. The value of losses in the implemented model tends to 0, which confirms the efficiency of its work. The value of box_loss is 0.01 to 500 iterations. This means that the objects are completely defined by the bounding box, i.e. the program finds objects entirely. Parameter obj_loss - loss reaches the value of 0.03 by 500 iterations. This shows that the program detects all objects, separating them from the background. The value of clc_loss - the loss of classification was 0 to 200 iterations, which indicates the correct distribution of objects by class.

Based on the result, we can conclude that the model shows good results in solving problems of detecting micro-objects in the marine environment.

As a result of research on the collected dataset of the training sample of photo and video images of microplankton and microplastics, the YOLOV5 model was trained, which solves the problem of automatic detection and recognition of micro-objects in the marine environment. Numerical quality metrics were selected and experimental studies were carried out for the implemented real-time model, which showed high reliability of the results obtained, comparable to manual recognition, when solving problems of detecting micro-objects in the marine environment.

**The study was supported by state assignment of Institute of natural and technical systems (Project Reg. No. 121122300070-9)**